\documentclass{article}
\usepackage{ijcai18}

\usepackage{times}
\usepackage{xcolor}
\usepackage{soul}
\usepackage[utf8]{inputenc}
\usepackage[small]{caption}
\usepackage{graphicx}

\title{Counting Cells in Time-Lapse Microscopy using  Deep Neural Networks}

\author{Alexander Gomez Villa$^1$, 
Augusto Salazar$^2$, 
Igor Stefanini$^3$ 
\\ 
$^{1,3}$ Istituto Sistemi e Electronica Applicata (ISEA) \\ Scuola universitaria professionale della Svizzera italiana (SUPSI) \\
$^2$ Grupo de investigacion SISTEMIC \\ Facultad de Ingenieria, Universidad de
Antioquia UdeA  \\
\{alexander.gomezvilla,igor.stefanini\}@supsi.ch,
augusto.salazar@udea.edu.co}

\begin{document}

\maketitle

\begin{abstract}
  An automatic approach to counting any kind of cells could alleviate work of the experts and boost the research in fields such as regenerative medicine. In this paper, a method for microscopy cell counting using multiple frames (hence temporal information) is proposed. Unlike previous approaches where the cell counting is done independently in each frame (static cell counting), in this work the cell counting prediction is done using multiple frames (dynamic cell counting). A spatiotemporal model using ConvNets and long short term memory (LSTM) recurrent neural networks is proposed to overcome temporal variations. The model outperforms static cell counting in a publicly available dataset of stem cells. The advantages, working conditions and limitations of the ConvNet-LSTM method are discussed. Although our method is tested in cell counting, it can be extrapolated to quantify in video (or correlated image series) any kind of objects or volumes.
\end{abstract}

\section{Introduction}


Analysis of cell images through time is a powerful tool commonly used in medicine and research. For instance, in regenerative medicine  scientists perform experiments in stem cells using several culture conditions and capture a large number of images to analyze them ~\cite{janmey2007cell}.
A particular case of time-lapse microscopy is the counting. Albeit simpler than other tasks (for instance cell segmentation) it would alleviate a lot of human effort in fields such as regenerative medicine. 

Automatic cell counting has been done with two approaches: detection and regression. In counting by detection the desired object to quantify is first detected, which involves a segmentation and classification step in traditional computer vision. This approach usually is designed for a specific cell type. Counting by detection has been replaced by more general counting techniques (counting by regression), borrowed from other computer vision problems (such as crown counting). On the other hand, counting by regression ignores the detection and estimates a number directly from the image. Hence, in this method usually the locations of the object can not be known. In this paper, we focus on the regression method. 

Several works have proposed approaches for counting cells by regression. Xie et. al.~\cite{xie2016microscopy} used a fully convolutional neural network to detect the position of the cells in the images. The input images were split into overlapping cuts that fit in the neural network. The full image was built again using interpolation over the neural network output. This approach allowed them to count using the integral over the output of the network. Xue et al.~\cite{xue2016cell} proposed using simple convolutional neural networks (ConvNet) as regressor to get the object count from the image. Using several mainstream ConvNets architectures and the same input image splitting as Xie et al.. Finally, Cohen et. al~\cite{cohen2017count} also use ConvNets as regressors in a pure regression way, but the image  in the final reconstruction stage changed. In his work each input pixel account for a number of cells, hence each image crop have redundancy counting (due to overlapping), which is used for total count prediction as a normalized factor below the total sum. 

These works have made big steps towards automation of cell counting, but they did not discuss the time factor. Although some microscopic objects do not change significantly in appearance over time (such as bacteria E.Coli), several kinds of cells transform their appearance over time (e.g. stem cells). Hence, a deeper analysis on how this automatic count approaches response to the time variation is needed.

In this work, the cell counting problem over time is faced. Several challenges related to the cell counting task in the single frame prediction (static) and multi-frame prediction (dynamic) cases are discussed. An evaluation and analysis of current approaches for cell counting in the dynamic context is done. Finally, a spatiotemporal model using ConvNets and long short term memory (LSTM) recurrent neural networks is proposed to overcome temporal variations. The advantages, working conditions and limitations of the ConvNet-LSTM method are discussed. Although our method is tested in cell counting it can be extrapolated to quantify in video (or correlated image series) any kind of objects or volumes. All annotations and derived datasets  used in this work, source code, and the trained models, are publicly available.

The rest of the paper is organized as follows: First, the challenges present in static and dynamic cell counting are discussed in section 2. Also, the methods used in the static and dynamic case are explained. Section 3 describes the experiments used to test the models. Results are presented in section 4. Section 5 discusses the results and limitations. Finally, in section 6 the conclusions and future work are presented.

\section{Methodology}

In this section different situations present in static and dynamic cases of cell counting are described and analyzed. The general ConvNet regression approach is briefly explained and the proposed spatiotemporal (ConvNet + LSTM) model is exposed.

\subsection{The counting Task}

\begin{figure}[hbtp]
	\centering	
	\includegraphics[width=0.35\textwidth]{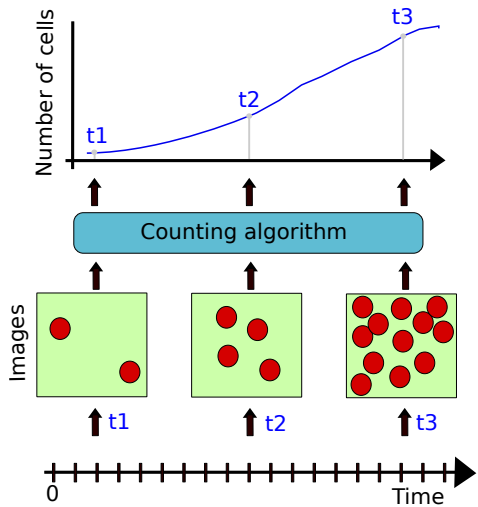}
	\caption{General procedure for cell counting}
	\label{fig:countingScene}
\end{figure}

The Fig.~\ref{fig:countingScene} shows the general procedure for cell counting. First, a cell culture or scene where the objects of interest are present is monitored over time. While the scene is observed images are taken periodically (sampled), this sampling frequency is parameter usually defined by the application (e.g. counting cells or counting people). The images (sampled in times t1, t2, and t3) are processed by a counting algorithm, which gives for each sampling time the number of cells (or objects) in the scene. In the Fig.~\ref{fig:countingScene} the interest objects (red dots) increase over time in the scene, but this condition is not mandatory.

\subsection{Challenges in static and dynamic cell counting}

The general perspective presented in Fig.~\ref{fig:countingScene} helps to state the problem, but several details remain behind.

\subsubsection{Background and Outliers}

The background (scene or context in which the objects of interest appear) usually is not uniform. In microscopy, several background noise contaminates the images, such as Non-uniform illumination, dead cells or organisms, external contamination agents (dust and particles), and scrap product of cells interactions and growing. This problem has been solved using a  sufficiently discriminative classifier in previous works of cell segmentation.

\subsubsection{Object appearance}

\begin{figure}[hbtp]
	\centering	
	\includegraphics[width=0.45\textwidth]{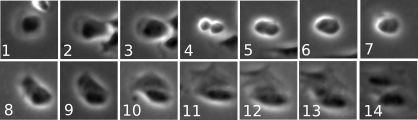}
	\caption{Snapshot of stem cell while reproduction takes place}
	\label{fig:cellRep}
\end{figure}

Cells are not time-invariant objects. They have a lot of intra-class variation caused for several reasons: change of nucleus position (if  visible), interactions between cells, internal mechanics of the cell, among others. Furthermore, the peak of intraclass variation occurs when the cell reproduction begins. Fig.~\ref{fig:cellRep} shows a set of frames of the same single cell while the reproduction takes place, the snapshots are sampled each hour. All the images (except the last one) from Fig.~\ref{fig:cellRep} must be identified as one cell despite the evident appearance variation. The number of possible variations increases when multiple cells inside the same image are considered (as each individual cell can be in any state of the reproduction independently). 


\subsubsection{Image partition}\label{sec:part}

\begin{figure}[hbtp]
	\centering	
	\includegraphics[width=0.35\textwidth]{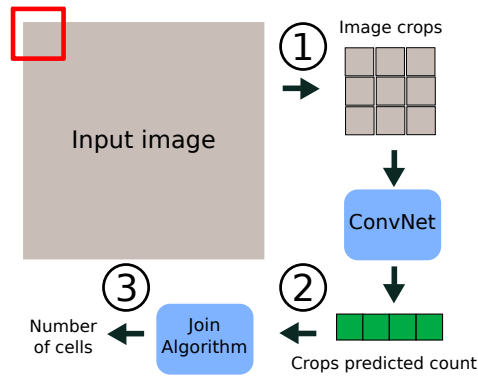}
	\caption{Static cell counting approach}
	\label{fig:cellparti}
\end{figure}

Current state-of-the-art proposed solutions~\cite{xie2016microscopy,xue2016cell} for cell counting has a similar framework to process the images. The Fig.~\ref{fig:cellparti} briefly reviews the approach.

The Fig.~\ref{fig:cellparti} shows the static cell counting approach. It can be summarized in three parts:

\begin{enumerate}
	\item The input image is cropped using a sliding window (red frame in Fig.~\ref{fig:cellparti}). The step and window size (usually equal to ConvNet input) must be selected a priori. Usually, the crops have an overlapping in order to have redundant information
	\item Each crop is evaluated in the ConvNet to get an estimation of the number of cells or objects
	\item An algorithm to merge the individual crops count information into a global count is used. Some works~\cite{xie2016microscopy,xue2016cell} use an interpolation to remake the input image as a density map (whose integral is the number of cells). Cohen et al.~\cite{cohen2017count}  sum all crops results and divide by the number of overlapping pixels.
\end{enumerate}

This approach has a known issue with the hyper-parameters (step and size of the sliding window) in step one. The size of the window controls the number of classes that the model must be able to classify. For instance, a window capable of containing four cells must classify between $0$,$1$,$2$,$3$, and $4$ cells (five classes), but a larger window (capable of containing eight cells) must classify nine classes. Hence, the window size controls the complexity of the classification task. A larger window will result in more classes to predict and fewer crops to train the model (since fewer partitions can be done). 

The step parameter controls the amount of redundancy and dataset size.  A little step will give a lot of images very correlated between them and higher step will result in more independent samples but also less training sets. Due to these reasons, size and step of the sliding window are problem-specific parameters i.e. its optimum values must be found experimentally  for each kind of cell (e.g. stem cells, blood cells, etc) or object. The theoretical relation between these parameters and the performance  of the model is an open question.

\subsubsection{Unbalanced data}

The number of samples per class (i.e the number of objects in each crop) could be unbalanced. For instance, the proliferation of cells can be quantified by the equation~\cite{sherley1995quantitative}:

\begin{equation}
N_t = N_0 2^{tf}
\end{equation}

where $N_t $ is the number of cells at time $t$, $N_0$ is the initial number of cells, and $f$ is the frequency of cell cycles per unit time. The exponential nature of this proliferation tells us that we are going to have far more images with few cells (e.g. $0$ and $1$). Hence, tiny datasets will suffer from a class unbalance problem which could limit the model performance. Big datasets do not suffer from unbalancing problems due to the classes can be artificially balanced erasing extra samples from most frequent classes.

\subsubsection{Sampling rate}

As Fig.~\ref{fig:countingScene} shows, the sampling times $t1$, $t2$, and $t3$ record an image of the scene to compute the number of cells. The sampling time has no effect in current counting approaches (due to time is not used in static cell counting), however in this work, the sampling time has influence on the features. A low sample rate (sample with less frequency) will lead to time-uncorrelated images, hence the time information can not be used. A high sample rate will have time-correlated images, but due to cells change (appearance and movement) in time very slow neighborhood images could be almost the same. 



\subsection{Long-term Recurrent Convolutional Network (LRCN) for dynamic cell counting }
 
\begin{figure}[hbtp]
	\centering	
	\includegraphics[width=0.35\textwidth]{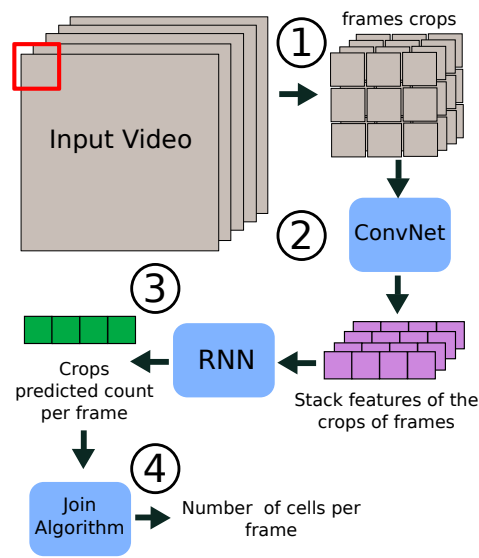}
	\caption{Proposed framework for cell counting in time-lapse microscopy}
	\label{fig:cellTime}
\end{figure}

The aim of our work is to merge the spatial data (static cell counting) with the time variable. This spatiotemporal problem has been previously addressed by the community of action recognition. Based on the work of Donahue et al.~\cite{donahue2015long} in action recognition, we propose a mix between ConvNets (to address images information) and recurrent neural networks (to deal with time Features). Fig.~\ref{fig:cellTime} shows the proposed framework for cell counting in time-lapse microscopy.

\subsubsection{Input data and partition}

In this case, the two parameters step and window size explained in section~\ref{sec:part} (image partition) remain equal. Additionally, a new parameter called temporal window ($tw$) is used. The $tw$ impose how many previous frames will be used to predict the current image number of objects. Albeit, this parameter ($tw$) adds a  constraint to the framework: The number of cells on the image can only be predicted when the number of previous frames is equal to $tw$. This condition must be avoided (especially in cell counting applications) since sample rate could be $1$ image per hour and the growing process usually takes several days. The method should be able to process the video frame by frame in order to take actions (or not) each time an image is recorded. In order to  avoid this constraint, a temporal padding is done, which will be explained in the following sections

Each frame is divided using the selected step and window size making a set or crops for each frame. Each crop has a set of temporally associated crops, i.e. the same crop (same spatial position) in past time instants (frames). The number of time-associated crops will be $tw$. Hence, as process number $1$ in Fig.\ref{fig:cellTime} shows, each frame is divided in crops and each crop is associated with $tw$ past crops.

\subsubsection{ConvNet}

The convolutional neural network acts as a feature extractor. Each crop is evaluated in the ConvNet, but instead of using the whole ConvNet as a regressor (as previous works did), we collect the feature vector that appears before the fully connected layer (the exact layer changes between architectures). These ConvNet features are usually extracted from pre-trained architectures and called off-the-shell or bottleneck features.

Four mainstream ConvNet architectures were used as feature extractors: VGG16, VGG19~\cite{simonyan2014very}, ResNet50~\cite{he2016deep}, and InceptionV3~\cite{szegedy2016rethinking}. Nevertheless, any ConvNet or computer vision approach can be used to make this mapping (image - features).

\subsubsection{Temporal padding and stack of features}

Once the crops are forwardly propagated through the ConvNet, a feature vector of dimension $1\times m$  is obtained for each crop ($m$ is the number of features, it depends on the ConvNet architecture e.g. $m=512$ for the VGG16). This feature vector must be stacked with its $tw$ previous crop's feature vector to build a block $\alpha$ of dimensions $tw\times m$. However, is no possible to  stack feature vectors when the number of previous frames is less than $tw$. Since no previous information exists, instead we add vectors of ones of dimension $1\times m$. For example, if $tw=5$ but at the moment only three images are recorded the block $\alpha$ is composed of three feature vectors with dimensions $1\times m$  and two vectors with ones of dimensions  $1\times m$. Experimentally, we found that this vector of one helps the network to predict the true number of cells since it gives an idea of which growing phase is the culture (this can also be a problem as will be discussed in limitations). Notice that $\alpha$ must be built in the strict temporal order in which the frames appear.  

\subsubsection{Recurrent Neural Network (RNN) and LSTM}

Recurrent neural networks have been successfully used for tasks with complex temporal dynamics such as speech recognition and text generation. Long short-term memory~\cite{hochreiter1997long} RNN extend the scalability of RNN, allowing them to be trained in large topologies without exploding gradients problems. In this work we use LSTM networks, future references to RNN must be considered as LSTM. The mix model ConvNet + LSTM will be addressed as LRCN.

RNN has several time inference methods, in this work, the many-to-one approach is used. In many-to-one inference, $k$ inputs enter to the RNN before a prediction is done. In this work $k=tw$, which means the input to the RNN in the step 3 (see Fig.\ref{fig:cellTime}) is going to be the block $\alpha$. Hence, we have an RNN with $tw$ time steps and one output corresponding to the estimated number of cells in the crop.

Bidirectional LSTM RNN~\cite{graves2005framewise} get the most of temporal information stepping through the input time steps in forward and backward directions. However, bidirectional LSTM RNN can be used only when the whole input (all the time steps) are available. In this case it is not a problem, since our block  $\alpha$ already has all the time steps. Bidirectional LSTM RNN are also used and compared against LSTM RNN.

\subsubsection{Join algorithm}

In step 4 (see Fig.\ref{fig:cellTime}) for each crop in the current frame there is an estimated cell count. In order to make a global decision on the number of cells in the current frame the same approach used in ~\cite{xue2016cell}(previously briefly explained in Image partition section) was done.

\section{Experimental setup}

In this section, the datasets and the experiments carried out in this work, are described. Additionally, an explanation of implementation details (such as libraries and architecture parameters, hardware, and optimization methods) is included.

\subsection{dataset}

In this work, the publicly available dataset Cell Image Analysis Archive ~\cite{kanade2011cell} was used. This dataset contains Myoblastic stem cells videos during the growth of the culture. It uses phase-contrast microscopy imaging acquiring images at a frequency of every 5 minutes over a course of approximately 3.5 days using a Zeiss Axiovert T135V microscope. Each image contains 1392$\times$ 1040 pixels with a resolution of 1.3m/pixel. Five sets of images (090325-F0009, 90303-F0002, 090318-F0007, and 090303-F0006) from different cultures were randomly selected (this is a very important condition since the RNN easily can memorize a single culture growing curve). Each set of images was sub-sampled to have one image per hour (this is the sampling rate parameter disused in section 2.2). Also, from each frame one quadrant (evenly dividing the whole image into four regions) is randomly selected. Each frame was manually annotated with a red dot in the center of a cell.

Following image partition section procedure, the parameters step and window size must be optimized for each type of cell. The window sizes $50$, $100$, and $200$ with overlapping (between crops) of  $25\%$, $50\%$, and $70\%$ were tested. Experimentally for our dataset, the window size 50 and overlapping of $50\%$ (hence step of $25$ pixels) worked better. The training and test set were built using a 5 fold cross-validation like method. In each fold, four images sets were used for training and the last set was used as  test. 



\subsection{Experiments}

\subsubsection{Static cell counting}

The five folds where tested using the framework proposed in Fig.~\ref{fig:cellparti}  with the features from the four pre-trained mainstream ConvNets. The ConvNets were pretrained models in the popular ImageNet dataset. Notice that as ~\cite{xue2016cell} we tried to train from scratch these architectures, however the accuracy decreased (also  ~\cite{xue2016cell} ). The fully connected layer of each model was fine-tuned for a regression task, due to the advantages (accuracy) of the regression over classification for cell counting were already shown by ~\cite{xue2016cell}. Experimentally we found that models perform better  training them  with an L1  loss function (the same conclusion of  ~\cite{xue2016cell,cohen2017count}).

\subsubsection{Dynamic cell counting}

The five folds were tested using the framework proposed in Fig.~\ref{fig:cellparti}  with the same features from ConvNets of Static cell counting experiments. A two-layer RNN with $30$ LSTM cells per layer was stacked at the end of each ConvNet (instead of the fully connected layer). This RNN was trained using an L2 loss function, notice that this is a different loss function from static cell counting framework. This harsh loss function helped to increase the accuracy using the unbalanced dataset in the dynamic cell counting framework (not the case for static cell counting).

The same set of experiments (four ConvNets + LSTM RNN ) where repeated changing parameter $tw$ between the values : $10$, $20$, $30$. This set of experiments were done using a bidirectional LSTM also with two layers of $30$ cells.

\subsection{Performance metrics}

We use the same performance metric (mean absolute error) of  ~\cite{xue2016cell,cohen2017count}  since it has been used by many object counting papers. 

\subsection{Implementation details}

All network optimization and testing is performed using an NVIDIA GeForce GTX 1080-ti GPU  and implemented using the Keras API~\cite{chollet2015keras} with a Tensorflow backend~\cite{tensorflow2016}.

\section{Results}

This section shows the results of the previously stated experiments split into two sections: static cell counting and dynamic cell counting results.

\subsection{Static cell counting}

Table~\ref{tab:Static} shows the results of static cell counting using ConvNets. The first column shows the ConvNet architecture followed by the  specific MAE and standard deviation in the test set of each fold. Notice the different performance with each fold, in almost all the experiments the fold $F3$ had the best performance and the folds $F1$ - $F5$ the worst.

\begin{table}[h]
	\centering
	\caption{Results of static cell counting experiments}
	\label{tab:Static}
	\begin{tabular}{cccccc}
		\hline
		Model       & F1                                                    & F2                                                    & F3                                                    & F4                                                    & F5                                                     \\
		\hline
		VGG16       & \begin{tabular}[c]{@{}c@{}}67.20\\ $\pm 35.0$\end{tabular} & \begin{tabular}[c]{@{}c@{}}57.02\\ $\pm$19.5\end{tabular} & \begin{tabular}[c]{@{}c@{}}37.71\\ $\pm$18.3\end{tabular} & \begin{tabular}[c]{@{}c@{}}23.59\\ $\pm$11.9\end{tabular} & \begin{tabular}[c]{@{}c@{}}64.28\\ $\pm$57.0\end{tabular} \\
		VGG19       & \begin{tabular}[c]{@{}c@{}}52.01\\ $\pm$27.8\end{tabular}  & \begin{tabular}[c]{@{}c@{}}29.12\\ $\pm$15.6\end{tabular}  & \begin{tabular}[c]{@{}c@{}}28.53\\ $\pm$21.9\end{tabular}  & \begin{tabular}[c]{@{}c@{}}13.23\\ $\pm$5.62\end{tabular}  & \begin{tabular}[c]{@{}c@{}}45.40\\ $\pm$35.3\end{tabular}   \\
		ResNet50    & \begin{tabular}[c]{@{}c@{}}\textbf{22.90}\\ $\pm$\textbf{18.5}\end{tabular}  & \begin{tabular}[c]{@{}c@{}}\textbf{22.32}\\ $\pm$\textbf{13.2}\end{tabular}  & \begin{tabular}[c]{@{}c@{}}\textbf{19.86}\\ $\pm$\textbf{14.6}\end{tabular}  & \begin{tabular}[c]{@{}c@{}}\textbf{6.63}\\ $\pm$\textbf{4.39}
			\end{tabular}   & \begin{tabular}[c]{@{}c@{}}\textbf{14.51}\\ $\pm$\textbf{9.22}\end{tabular}   \\
		InceptionV3 & \begin{tabular}[c]{@{}c@{}}36.31\\ $\pm$27.2\end{tabular}  & \begin{tabular}[c]{@{}c@{}}26.3\\ $\pm$17.5\end{tabular}   & \begin{tabular}[c]{@{}c@{}}35.86\\ $\pm$29.3\end{tabular}  & \begin{tabular}[c]{@{}c@{}}27.62\\ $\pm$19.3\end{tabular}  & \begin{tabular}[c]{@{}c@{}}77.56\\ $\pm$62.7\end{tabular}  \\
		\hline
	\end{tabular}
\end{table}

The best architecture (in performance) was the ResNet50 and the worst the VGG16. Similar to ~\cite{xue2016cell} we got better results (in almost all results) with deeper architectures  using fine-tuning. The results with the pre-trained InceptionV3 ConvNet  could be due to the fact that the model is  too much specialized in natural images (ImageNet images), which are very different from microscopy images.

\subsection{Dynamic cell counting with LSTM}

Table~\ref{tab:LSTM} shows the results of dynamic cell counting using ConvNets and LSTM. The first column shows the ConvNet architecture followed by the specific MAE and standard deviation in the test set for each fold. Table~\ref{tab:LSTM} is subdivided by variation in the $tw$ parameter from $5$ to $30$ (as previously stated).

\begin{table}[ht]
	\centering
	\caption{Results of dynamic cell counting using LSTM experiments}
	\label{tab:LSTM}
	\begin{tabular}{cccccc}
		\hline
		\multicolumn{6}{c}{LRCN - $tw=10$}                                                                                                                                                                                                                                                                   \\
		\hline
		VGG16       & \begin{tabular}[c]{@{}c@{}}43.68\\ $\pm$36.7\end{tabular} & \begin{tabular}[c]{@{}c@{}}10.16\\ $\pm$6.9\end{tabular}  & \begin{tabular}[c]{@{}c@{}}12.96\\ $\pm$5.63\end{tabular} & \begin{tabular}[c]{@{}c@{}}51.50\\ $\pm$18.3\end{tabular}  & \begin{tabular}[c]{@{}c@{}}55.17\\ $\pm$50.9\end{tabular} \\
		VGG19       & \begin{tabular}[c]{@{}c@{}}32.21\\ $\pm$25.7\end{tabular} & \begin{tabular}[c]{@{}c@{}}9.99\\ $\pm$5.5\end{tabular}   & \begin{tabular}[c]{@{}c@{}}9.83\\ $\pm$5.79\end{tabular} & \begin{tabular}[c]{@{}c@{}}29.14\\ $\pm$20.8\end{tabular}  & \begin{tabular}[c]{@{}c@{}}29.14\\ $\pm$20.9\end{tabular} \\
		ResNet50    & \begin{tabular}[c]{@{}c@{}}42.42\\ $\pm$37.0\end{tabular} & \begin{tabular}[c]{@{}c@{}}43.36\\ $\pm$14.0\end{tabular} & \begin{tabular}[c]{@{}c@{}}28.25\\ $\pm$13.2\end{tabular} & \begin{tabular}[c]{@{}c@{}}23.13\\ $\pm$11.6\end{tabular}  & \begin{tabular}[c]{@{}c@{}}56.37\\ $\pm$53.4\end{tabular} \\
		InceptionV3 & \begin{tabular}[c]{@{}c@{}}53.25\\ $\pm$43.8\end{tabular} & \begin{tabular}[c]{@{}c@{}}22.96\\ $\pm$12.5\end{tabular} & \begin{tabular}[c]{@{}c@{}}19.37\\ $\pm$10.4\end{tabular} & \begin{tabular}[c]{@{}c@{}}24.03\\ $\pm$12.1\end{tabular}  & \begin{tabular}[c]{@{}c@{}}58.89\\ $\pm$57.5\end{tabular} \\
		\hline
		\multicolumn{6}{c}{LRCN - $tw=20$}                                                                                                                                                                                                                                                                   \\
		\hline
		VGG16       & \begin{tabular}[c]{@{}c@{}}42.13\\ $\pm$34.2\end{tabular} & \begin{tabular}[c]{@{}c@{}}10.95\\ $\pm$7.29\end{tabular} & \begin{tabular}[c]{@{}c@{}}33.89\\ $\pm$5.75\end{tabular} & \begin{tabular}[c]{@{}c@{}}36.38\\ $\pm$16.9\end{tabular}  & \begin{tabular}[c]{@{}c@{}}30.19\\ $\pm$25.6\end{tabular} \\
		VGG19       & \begin{tabular}[c]{@{}c@{}}\textbf{30.38}\\ $\pm$\textbf{26.3}\end{tabular} & \begin{tabular}[c]{@{}c@{}}\textbf{7.26}\\ $\pm$\textbf{4.5}\end{tabular}   & \begin{tabular}[c]{@{}c@{}}\textbf{8.63}\\ $\pm$\textbf{5.76}\end{tabular} & \begin{tabular}[c]{@{}c@{}}\textbf{35.82}\\ $\pm$\textbf{14.7}\end{tabular}  & \begin{tabular}[c]{@{}c@{}}\textbf{27.63}\\ $\pm$\textbf{19.5}\end{tabular} \\
		ResNet50    & \begin{tabular}[c]{@{}c@{}}42.94\\ $\pm$37.6\end{tabular} & \begin{tabular}[c]{@{}c@{}}18.55\\ $\pm$10.4\end{tabular} & \begin{tabular}[c]{@{}c@{}}25.80\\ $\pm$10.9\end{tabular} & \begin{tabular}[c]{@{}c@{}}22.34\\ $\pm$11.8\end{tabular}  & \begin{tabular}[c]{@{}c@{}}54.16\\ $\pm$47.8\end{tabular} \\
		InceptionV3 & \begin{tabular}[c]{@{}c@{}}42.57\\ $\pm$37.1\end{tabular} & \begin{tabular}[c]{@{}c@{}}11.45\\ $\pm$9.12\end{tabular} & \begin{tabular}[c]{@{}c@{}}17.03\\ $\pm$13.4\end{tabular} & \begin{tabular}[c]{@{}c@{}}21.10\\ $\pm$11.0\end{tabular}  & \begin{tabular}[c]{@{}c@{}}52.68\\ $\pm$33.0\end{tabular} \\
		\hline
		\multicolumn{6}{c}{LRCN - $tw=30$}                                                                                                                                                                                                                                                                   \\
		\hline
		VGG16       & \begin{tabular}[c]{@{}c@{}}34.81\\ $\pm$30.2\end{tabular} & \begin{tabular}[c]{@{}c@{}}25.60\\ $\pm$13.3\end{tabular} & \begin{tabular}[c]{@{}c@{}}31.66\\ $\pm$16.0\end{tabular} & \begin{tabular}[c]{@{}c@{}}27.49\\ $\pm$16.0\end{tabular}  & \begin{tabular}[c]{@{}c@{}}53.94\\ $\pm$47.1\end{tabular} \\
		VGG19       & \begin{tabular}[c]{@{}c@{}}46.93\\ $\pm$34.5\end{tabular} & \begin{tabular}[c]{@{}c@{}}13.23\\ $\pm$6.48\end{tabular} & \begin{tabular}[c]{@{}c@{}}19.87\\ $\pm$6.45\end{tabular} & \begin{tabular}[c]{@{}c@{}}30.49\\ $\pm$19.6\end{tabular}  & \begin{tabular}[c]{@{}c@{}}53.87\\ $\pm$46.9\end{tabular} \\
		ResNet50    & \begin{tabular}[c]{@{}c@{}}45.48\\ $\pm$40.1\end{tabular} & \begin{tabular}[c]{@{}c@{}}19.36\\ $\pm$11.0\end{tabular} & \begin{tabular}[c]{@{}c@{}}29.03\\ $\pm$13.9\end{tabular} & \begin{tabular}[c]{@{}c@{}}24.41\\ $\pm$12.5\end{tabular}  & \begin{tabular}[c]{@{}c@{}}54.74\\ $\pm$49.7\end{tabular} \\
		InceptionV3 & \begin{tabular}[c]{@{}c@{}}42.15\\ $\pm$36.7\end{tabular} & \begin{tabular}[c]{@{}c@{}}30.79\\ $\pm$14.9\end{tabular} & \begin{tabular}[c]{@{}c@{}}29.31\\ $\pm$14.2\end{tabular} & \begin{tabular}[c]{@{}c@{}}19.5\\ $\pm$18.7\end{tabular}   & \begin{tabular}[c]{@{}c@{}}53.65\\ $\pm$46.3\end{tabular} \\
		\hline
	\end{tabular}
\end{table}


Notice that the unbalance performance between folds continues: The fold $F3$ had the best performance and the folds $F1$ - $F5$ the worst (for all the values of $tw$). When $tw$ increases the performance has a mostly regular increasing pattern until $tw=30$. As Table~\ref{tab:LSTM} in $tw=30$ shows, too much temporal information (high $tw$) reduces the performance  and increase the standard deviation. However, for $tw=20$ (the best value experimentally found for $tw$) a direct comparison between VGG16 and VGG19 architectures (e.g. VGG16 results in static cell counting against VGG16 in dynamic cell counting) shows how LRCN is better in almost all the cases than single ConvNet.

Nevertheless, for the complex ConvNets (residual connections and network in network in the full connected layer) the coupling approach did not perform well. This result is probably due to the full connected layer in the ConvNets perform better than our stacked two-layer LSTM network used in the LRCN. A more specific and complex LSTM network probably could beat single ConvNet approach in the  ResNet and InceptionV3 cases (however this hypothesis is not proved in this paper).

Table~\ref{tab:BiLSTM} shows the results of dynamic cell counting using ConvNets and bidirectional LSTM (Bi-LSTM). The distribution of the results in the table is the same as Table~\ref{tab:LSTM}. As expected the Bi-LRCN perform better (in almost all the cases) than the simple LRCN. Following the results of ~\cite{graves2005framewise} the information of the sequence in backward direction has a great impact on the model. Notice the Bi-LRCN improve  the MAE but not so much the standard deviation.

\begin{table}[ht]
	\centering
	\caption{Results of dynamic cell counting using Bi-LSTM}
	\label{tab:BiLSTM}
	\begin{tabular}{cccccc}
		\hline
		\multicolumn{6}{c}{Bi-LRCN - $tw=10$}                                                                                                                                                                                                                                                                \\
		\hline
		VGG16       & \begin{tabular}[c]{@{}c@{}}40.02\\ $\pm$33.0\end{tabular} & \begin{tabular}[c]{@{}c@{}}12.04\\ $\pm$9.32\end{tabular} & \begin{tabular}[c]{@{}c@{}}12.33\\ $\pm$5.46\end{tabular}  & \begin{tabular}[c]{@{}c@{}}36.65\\ $\pm$17.8\end{tabular} & \begin{tabular}[c]{@{}c@{}}25.30\\ $\pm$16.3\end{tabular} \\
		VGG19       & \begin{tabular}[c]{@{}c@{}}28.16\\ $\pm$26.0\end{tabular} & \begin{tabular}[c]{@{}c@{}}11.69\\ $\pm$5.93\end{tabular} & \begin{tabular}[c]{@{}c@{}}28.31\\ $\pm$6.50\end{tabular}  & \begin{tabular}[c]{@{}c@{}}25.84\\ $\pm$13.7\end{tabular} & \begin{tabular}[c]{@{}c@{}}27.87\\ $\pm$21.3\end{tabular} \\
		ResNet50    & \begin{tabular}[c]{@{}c@{}}43.71\\ $\pm$38.5\end{tabular} & \begin{tabular}[c]{@{}c@{}}17.42\\ $\pm$9.39\end{tabular} & \begin{tabular}[c]{@{}c@{}}23.94\\ $\pm$9.37\end{tabular}  & \begin{tabular}[c]{@{}c@{}}25.18\\ $\pm$13.1\end{tabular} & \begin{tabular}[c]{@{}c@{}}54.14\\ $\pm$44.7\end{tabular} \\
		nceptionV3 & \begin{tabular}[c]{@{}c@{}}50.54\\ $\pm$21.5\end{tabular} & \begin{tabular}[c]{@{}c@{}}12.11\\ $\pm$10.2\end{tabular} & \begin{tabular}[c]{@{}c@{}}18.12\\ $\pm$12.2\end{tabular}   & \begin{tabular}[c]{@{}c@{}}18.71\\ $\pm$15.6\end{tabular} & \begin{tabular}[c]{@{}c@{}}58.92\\ $\pm$40.2\end{tabular} \\
		\hline
		\multicolumn{6}{c}{Bi-LRCN - $tw=20$}                                                                                                                                                                                                                                                                \\
		\hline
		VGG16       & \begin{tabular}[c]{@{}c@{}}38.03\\ $\pm$31.5\end{tabular} & \begin{tabular}[c]{@{}c@{}}8.88\\ $\pm$6.88\end{tabular}  & \begin{tabular}[c]{@{}c@{}}9.79\\ $\pm$ 5.61\end{tabular} & \begin{tabular}[c]{@{}c@{}}33.47\\ $\pm$16.7\end{tabular} & \begin{tabular}[c]{@{}c@{}}28.63\\ $\pm$25.8\end{tabular} \\
		VGG19       & \begin{tabular}[c]{@{}c@{}}\textbf{27.42}\\ $\pm$\textbf{23.1}\end{tabular} & \begin{tabular}[c]{@{}c@{}}\textbf{5.43}\\ $\pm$\textbf{4.9}\end{tabular}   & \begin{tabular}[c]{@{}c@{}}\textbf{7.87}\\ $\pm$\textbf{5.67}\end{tabular}  & \begin{tabular}[c]{@{}c@{}}\textbf{43.23}\\ $\pm$\textbf{15.4}\end{tabular} & \begin{tabular}[c]{@{}c@{}}\textbf{25.43}\\ $\pm$\textbf{18.1}\end{tabular} \\
		ResNet50    & \begin{tabular}[c]{@{}c@{}}45.52\\ $\pm$38.5\end{tabular} & \begin{tabular}[c]{@{}c@{}}19.21\\ $\pm$12.42\end{tabular} & \begin{tabular}[c]{@{}c@{}}25.63\\ $\pm$9.1\end{tabular}  & \begin{tabular}[c]{@{}c@{}}26.54\\ $\pm$14.2\end{tabular} & \begin{tabular}[c]{@{}c@{}}54.33\\ $\pm$50.4\end{tabular} \\
		InceptionV3 & \begin{tabular}[c]{@{}c@{}}45.15\\ $\pm$21.5\end{tabular} & \begin{tabular}[c]{@{}c@{}}25.03\\ $\pm$14.0\end{tabular} & \begin{tabular}[c]{@{}c@{}}23.12\\ $\pm$8.9\end{tabular}   & \begin{tabular}[c]{@{}c@{}}20.90\\ $\pm$12.2\end{tabular} & \begin{tabular}[c]{@{}c@{}}59.83\\ $\pm$29.2\end{tabular} \\
		\hline
		\multicolumn{6}{c}{Bi-LRCN - $tw=30$}                                                                                                                                                                                                                                                                \\
		\hline
		VGG16       & \begin{tabular}[c]{@{}c@{}}40.41\\ $\pm$33.1\end{tabular} & \begin{tabular}[c]{@{}c@{}}10.27\\ $\pm$7.29\end{tabular} & \begin{tabular}[c]{@{}c@{}}23.28\\ $\pm$7.19\end{tabular}  & \begin{tabular}[c]{@{}c@{}}66.37\\ $\pm$19.0\end{tabular} & \begin{tabular}[c]{@{}c@{}}27.78\\ $\pm$22.4\end{tabular} \\
		VGG19       & \begin{tabular}[c]{@{}c@{}}24.86\\ $\pm$21.5\end{tabular} & \begin{tabular}[c]{@{}c@{}}6.49\\ $\pm$4.2\end{tabular}   & \begin{tabular}[c]{@{}c@{}}21.49\\ $\pm$5.71\end{tabular}  & \begin{tabular}[c]{@{}c@{}}29.20\\ $\pm$13.8\end{tabular} & \begin{tabular}[c]{@{}c@{}}24.33\\ $\pm$21.3\end{tabular} \\
		ResNet50    & \begin{tabular}[c]{@{}c@{}}53.44\\ $\pm$43.8\end{tabular} & \begin{tabular}[c]{@{}c@{}}23.59\\ $\pm$12.7\end{tabular} & \begin{tabular}[c]{@{}c@{}}27.22\\ $\pm$12.3\end{tabular}  & \begin{tabular}[c]{@{}c@{}}22.28\\ $\pm$11.3\end{tabular} & \begin{tabular}[c]{@{}c@{}}54.74\\ $\pm$49.7\end{tabular} \\
		InceptionV3 & \begin{tabular}[c]{@{}c@{}}44.43\\ $\pm$23.7\end{tabular} & \begin{tabular}[c]{@{}c@{}}24.56\\ $\pm$14.0\end{tabular} & \begin{tabular}[c]{@{}c@{}}22.07\\ $\pm$9.3\end{tabular}   & \begin{tabular}[c]{@{}c@{}}18.92\\ $\pm$12.8\end{tabular} & \begin{tabular}[c]{@{}c@{}}61.90\\ $\pm$60.5\end{tabular} \\
		\hline
	\end{tabular}
\end{table}

\section{Discussion and Limitations}

Although dynamic cell counting was shown to overcome static cell counting, the following issues were found to limit the performance of our approach:

\subsection{Unsuccesfull improvement in LRCN using ResNet and InceptionV3}

As was previously stated, simple (no residual connections or network in network approaches) Convnets such as VGG16 and VGG19 the LRCN improve a lot the performance, but for complex ConvNets the approach did not perform well. Complex LSTM networks, which can replace the full connected layer of these ConvNets, must be implemented.

\subsection{Unbalanced performance between folds}

The performance ranking between fold remains more or less equal in all the experiments, from high to less MAE: $F5-F1$ followed by $F4-F2-F3$. These results reflect the number of samples per class in each test set in the folds. $F5$ and $F1$ are the sets with a higher number of classes (from 0 to 8 cells per images). Therefore the model did not have enough samples from classes with a lot (more than four) of cells and perform worst in this test sets. However, the fact that the LRCN deals better with this problem proves the positive influence of temporal information.

\subsection{LRCN train set issues}

In ~\cite{xue2016cell,cohen2017count} some ConvNets are trained from scratch leading to better results than fine-tuned ones. In our case this task was impossible  for both models: static cell counting and dynamic cell counting giving always worst results than fine-tuned models. This problem has two cores: Dataset size and unbalancing. We proceeded to balance the folds leaving a maximum of $1000$ samples per class and leaving low number classes the same. Table~\ref{tab:StaticBala} shows the results of static cell counting with ``balanced'' folds.

\begin{table}[h]
	\centering
	\caption{Results of static cell counting with balanced folds}
	\label{tab:StaticBala}
	\begin{tabular}{cccccc}
		\hline
		Model       & F1                                                    & F2                                                    & F3                                                    & F4                                                    & F5                                                     \\
		\hline
		VGG16       & \begin{tabular}[c]{@{}c@{}}21.77\\ $\pm$14.3\end{tabular} & \begin{tabular}[c]{@{}c@{}}54.84\\ $\pm$11.6\end{tabular} & \begin{tabular}[c]{@{}c@{}}66.44\\ $\pm$5.9\end{tabular} & \begin{tabular}[c]{@{}c@{}}32.03\\ $\pm$14.7\end{tabular} & \begin{tabular}[c]{@{}c@{}}54.65\\ $\pm$10.9\end{tabular} \\
		VGG19       & \begin{tabular}[c]{@{}c@{}}17.3\\ $\pm$8.6\end{tabular}  & \begin{tabular}[c]{@{}c@{}}15.11\\ $\pm$12.8\end{tabular}  & \begin{tabular}[c]{@{}c@{}}9.64\\ +10.1\end{tabular}  & \begin{tabular}[c]{@{}c@{}}27.74\\ $\pm$20.9\end{tabular}  & \begin{tabular}[c]{@{}c@{}}16.04\\ $\pm$10.3\end{tabular}   \\
		ResNet50    & \begin{tabular}[c]{@{}c@{}}6.45\\ $\pm$5.58\end{tabular}  & \begin{tabular}[c]{@{}c@{}}14.58\\ $\pm$14.1\end{tabular}  & \begin{tabular}[c]{@{}c@{}}6.43\\ $\pm$6.2\end{tabular}  & \begin{tabular}[c]{@{}c@{}}27.42\\ $\pm$9.5\end{tabular}   & \begin{tabular}[c]{@{}c@{}}22.55\\ $\pm$9.6\end{tabular}   \\
		InceptionV3 & \begin{tabular}[c]{@{}c@{}}25.43\\ $\pm$13.2\end{tabular}  & \begin{tabular}[c]{@{}c@{}}14.07\\ $\pm$8.8\end{tabular}   & \begin{tabular}[c]{@{}c@{}}46.88\\ $\pm$22.6\end{tabular}  & \begin{tabular}[c]{@{}c@{}}38.79\\ $\pm$7.0\end{tabular}  & \begin{tabular}[c]{@{}c@{}}26.95\\ $\pm$18.2\end{tabular}  \\
		\hline
	\end{tabular}
\end{table}

Table~\ref{tab:StaticBala} shows a lot of improvement with respect to unbalanced static cell counting. However, this same approach could not be done with LRCN, due to the reduced train set the model memorizes the dataset very fast, avoiding any generalization. The authors believe that with a larger dataset a balanced version of the database could lead to even better performance using the LRCN model.

\section{Conclusions}

In this paper, a computer vision cell counting approach using multiple frames (hence temporal information) is proposed. An extension of previous works using a mixed architecture with convolutional neural networks (ConvNets) and recurrent neural networks is developed. Using a publicly available dataset of stem cells and four mainstream deep ConvNets a comparison between frame cell counting prediction (static cell counting) and proposed multi-frame cell counting (dynamic cell counting) was shown. The results show how dynamic cell counting surpasses the static cell counting approach and resists better the unbalancing nature of microscopy image data. A detailed analysis of challenges and common issues in cell counting for time-lapse microscopy was also presented.

In the future, several architectures must be tested. A specific ConvNet architecture as ~\cite{cohen2017count} proposed could lead to better results. Also, $3$ dimensional ConvNets (input frames overlapped as layers) and attention based models (in recurrent neural networks) could enhance performance. However, the most important issue is the data size. 

\section*{Acknowledgments}

We  would like to thank RETECA Foundation for the support and the opportunity to develop this work.

\bibliographystyle{named}
\bibliography{ijcai18}

\end{document}